\documentclass[conference,a4paper]{IEEEtran}
\ifCLASSINFOpdf
\else
\fi

\usepackage{amsmath}
\usepackage{cite}
\usepackage{graphicx}
\usepackage{comment}
\usepackage{adjustbox}
\usepackage{amsfonts}
\usepackage{multirow}
\usepackage{makecell}
\usepackage{bigstrut}
\usepackage{epstopdf}
\usepackage[hyphens]{url}
\usepackage[hidelinks]{hyperref}
\hypersetup{breaklinks=true}
\begin{document}
%\title{Motion Assisted Fall Detection System based on Skeleton Joints}
\title{Modeling Human Skeleton Joint Dynamics for \\ Fall Detection}

%we propose a skeleton based method that exploits spatio-temporal joint dependency and dynamic motion for efficient fall detection.

\author{\IEEEauthorblockN{Sania Zahan}
\IEEEauthorblockA{The University of Western Australia\\
Perth, Australia\\
sania.zahan@research.uwa.edu.au}
\and
\IEEEauthorblockN{Ghulam Mubashar Hassan}
\IEEEauthorblockA{The University of Western Australia\\
Perth, Australia\\
ghulam.hassan@uwa.edu.au}
\and
\IEEEauthorblockN{Ajmal Mian}
\IEEEauthorblockA{The University of Western Australia\\
Perth, Australia\\
ajmal.mian@uwa.edu.au}}


%Telephone: (800) 555--1212, Fax: (888) 555--1212}


\maketitle

\begin{abstract}
%\boldmath
The increasing pace of population aging calls for better care and support systems. Falling is a frequent and critical problem for elderly people causing serious long-term health issues. Fall detection from video streams is not an attractive option for real-life applications due to privacy issues. Existing methods try to resolve this issue by using very low-resolution cameras or video encryption. However, privacy cannot be ensured completely with such approaches. Key points on the body, such as skeleton joints, can convey significant information about motion dynamics and successive posture changes which are crucial for fall detection. Skeleton joints have been explored for feature extraction but with image recognition models that ignore joint dependency across frames which is important for the classification of actions. Moreover, existing models are over-parameterized or evaluated on small datasets with very few activity classes. We propose an efficient graph convolution network model that exploits spatio-temporal joint dependencies and dynamics of human skeleton joints for accurate fall detection. Our method leverages dynamic representation with robust concurrent spatio-temporal characteristics of skeleton joints. We performed extensive experiments on three large-scale datasets. With a significantly smaller model size than most existing methods, our proposed method achieves state-of-the-art results on the large scale NTU datasets. %Ajmal: you are not using RGB or depth
\end{abstract}
% \fi
\IEEEpeerreviewmaketitle
 
\section{Introduction}
Falling occurs commonly in our daily life and pose a major public health problem globally. It is especially dangerous for elderly people and children who are more susceptible to injuries. According to the World Health Organisation (WHO), it is the second leading cause of unintentional injury based deaths worldwide after road traffic accidents \cite{WHO_FALL}. %Though not as severe, young people also suffer considerably. 
Similarly, WHO statistics for children in China show that for every death due to a fall, there are 4 cases of permanent disability, 13 cases requiring hospitalization for over 10 days, 24 cases of hospitalization for 1–9 days, and 690 cases seeking medical care or missing work/school \cite{WHO_FALL}. 

\begin{figure}[htbp]
\centering
\includegraphics[height=5.5cm, width=8cm]{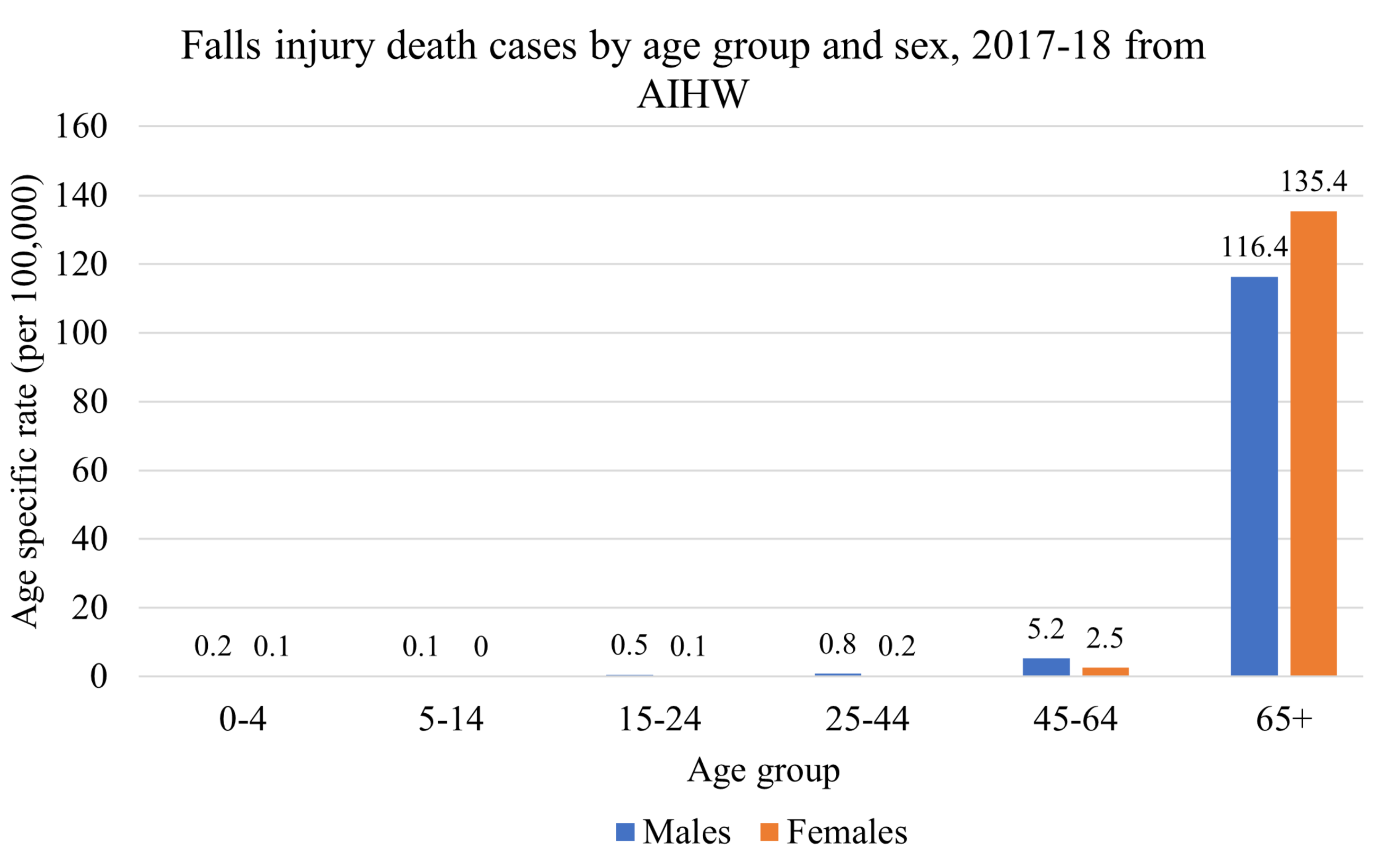}
\caption{Falls injury death cases, by age group and sex, 2017–18 from AIHW National Hospital Morbidity Database \cite{AIHW_FALL}.} %People aged over 65 have a much higher chance of death from fall.}
\label{fig:death_cases}
\end{figure}

The fall problem becomes more critical for people living alone with no immediate access to assistance. People aged over 65 years are more prone to falls due to deteriorating health and balance problems. %Neurodegenerative diseases like Mild cognitive impairment (MCI), a common issue in old age, can cause dementia, loss of visual/spatial perception resulting in gait/balance disorders \cite{Romeo_vision}. Besides, people with epilepsy, blood sugar issues etc. can suddenly lose control over muscles and fall. 
They are more likely to suffer from moderate to severe injuries such as bone fractures or dislocation, head trauma etc. Recovery is slow among them causing long-term post-operative complications and may also result in complete immobility. Each year 37.3 million falls occur worldwide that are severe enough to require medical attention \cite{WHO_FALL}. According to the Australian Institute of Health and Welfare (AIHW) National Hospital Morbidity Database, 5156 deaths occurred from more than 2 million hospitalization cases in the year 2017-18 in Australia due to falls \cite{AIHW_FALL}. Among those cases, 1.7\% of patients were admitted to intensive care units with 0.4\% needing continuous ventilator support. Fig.~\ref{fig:death_cases} shows total death cases from fall injuries in 2017-18 indicating an even higher risk for females aged over 65. 

\begin{figure}[htbp]
\centering
\includegraphics[height=7cm, width=8.9cm]{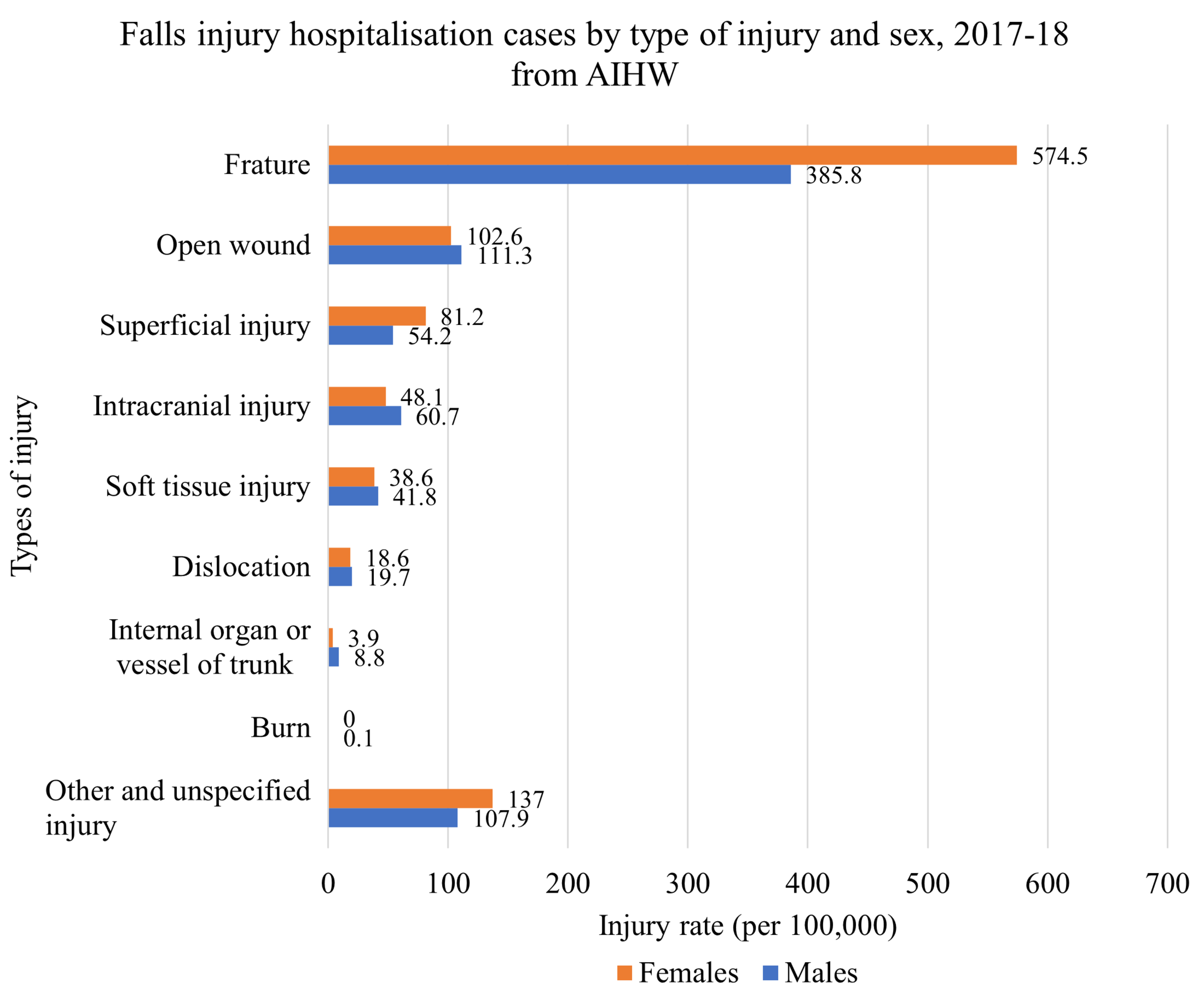}
\caption{Fall related injury in hospitalisation cases, by type of injury and sex, 2017–18 from AIHW National Hospital Morbidity Database \cite{AIHW_FALL}.} %Especially women are at higher risk from injuries like fracture.}
\label{fig:hospitalization_cases}
\end{figure}

%Head and neck are the prominent body parts that suffer from an injury that may result in long-term difficulties. 
Fig.~\ref{fig:hospitalization_cases} illustrates rates of different types of injuries due to falls in hospitalization cases per million in 2017-18. Fractures with the highest rate was the most common injury. Other injuries like open wounds, superficial and intracranial injuries were also significant. Many of these injuries can result in permanent disability.

In 2010, 14\% of the total population in Australia was over 65 years old. It is predicted to increase up to 23\% by 2050 \cite{ANZ_FALL}. Similarly, the UN report on the aging population suggests that the global population aged over 60 is going to exceed 2 billion by 2050~\cite{UN_age}. This upcoming elderly society will require efficient and large health care support. In Australia only, medical cost due to falls is expected to increase to \$1.4 billion by 2051 \cite{ANZ_FALL}. This cost does not even consider the decreased quality of life and long-term effects. 

Another consequence of falling is ``long lie" where the person remains on the ground for more than an hour after a fall possibly due to weakness or unconsciousness. This mostly happens with people living in social isolation causing a delay in treatment. Studies have found an association of ``long lie" to increased risk of physical and physiological complications \cite{long_lie_jama}. In addition to physical difficulties, elder people may also suffer from ``post-fall syndrome” losing confidence in independent mobility \cite{MATHON2017e50}.% thus increasing the probability of being admitted to an institution. 

Given the severity of fall related consequences, it is crucial to ensure proper monitoring and care facilities for elderly people to improve their quality of life and safety. An effective automated alert system can assist in continuous monitoring with prompt alerts to caregivers and emergency services. This will significantly reduce delay in assistance and facilitate better access to care. 

Many efforts have already been made to develop and improve the accuracy of fall detection and prediction systems. Traditional fall detection systems use various  wearable sensory data to detect falls \cite{NYAN_2008, Fakhrulddin_2017, NGUYENGIA_2018, Alarifi_2021, Anceschi_2021}. A disadvantage of these methods is that a person needs to wear the device at all times which is normally inconvenient. Besides, these techniques differentiate fall from other actions by applying a threshold that detects the rate of change in motion. Thus, such systems suffer from high false alarms due to the variety of daily activities like slapping a hand on the table, clapping with both hands, shaking sand out of a blanket, or even chopping vegetables. The optimal choice for a threshold is quite challenging and varies from person to person. Depending on physical and movement characteristics, for some people, these devices will produce more false alarms and may miss actual falls for others. %To compensate for these variations, some systems approach fusing signals from different sensors. But this leads to a costly solution. 

Recently, vision based fall detection methods \cite{Lin_lstm,  Keskes_2021,  Carlier_2020_EMBC,  Wang_2020, Liu_ICAIBD, FENG2020242,  XU2020123205,  Cai_AE_2020} are found to be better at differentiating diverse types of daily life activities. They mostly use RGB video feed which is not suitable in all situations due to privacy issues. Therefore, we need to consider these aspects to design an effective and robust fall detection system. 

Humans are capable of identifying different types of actions simply from biological motions without any appearance information \cite{Gunnar_1973}. This phenomenon invoked an immense interest in skeleton joint based human action analysis in recent years. Skeleton joints can capture rich spatio-temporal motion dynamics and are also computationally efficient due to their low dimensional representation. They are also robust to changes in the viewpoint, lighting conditions and background etc. There has been very little research on skeleton based fall detection and most solutions are based on image processing models. However, treating skeleton as a graph is much more intuitive than as an image considering the natural node-edge structure of joints and bones. Therefore, in this work, we propose a skeleton based method that exploits spatio-temporal joint dependency and dynamic motion for efficient fall detection. We tested our method on three publicly available datasets including the large scale NTU 120 dataset \cite{Liu_2019_NTURGBD120} that contains 120 different types of daily activities.

The rest of the paper is organized as follows: Section~\ref{Related_Works} analyzes related works, Section~\ref{method} describes the proposed fall detection method, Section~\ref{results} represents the experimental results and comparison with state-of-the-art results, and finally, conclusions are drawn in Section~\ref{conclusion}.
%-------------------------------------------------------------------------

\section{Related Work}\label{Related_Works}
We can categorise current works on fall detection methods into three categories: wearable, environmental and vision-based sensors. Wearable sensors mostly use multi-axis accelerometers to detect accelerations, velocities and angles of the wearer and use simple thresholding to identify falls. Environmental sensors, on the other hand, detect falls through vibration, infrared sensors or audio. Vision-based methods are the most promising as it is easier and more effective to be incorporated into today's smart-home systems. 

\begin{figure}[htbp]
\centering
\includegraphics[height=5cm, width=6.5cm]{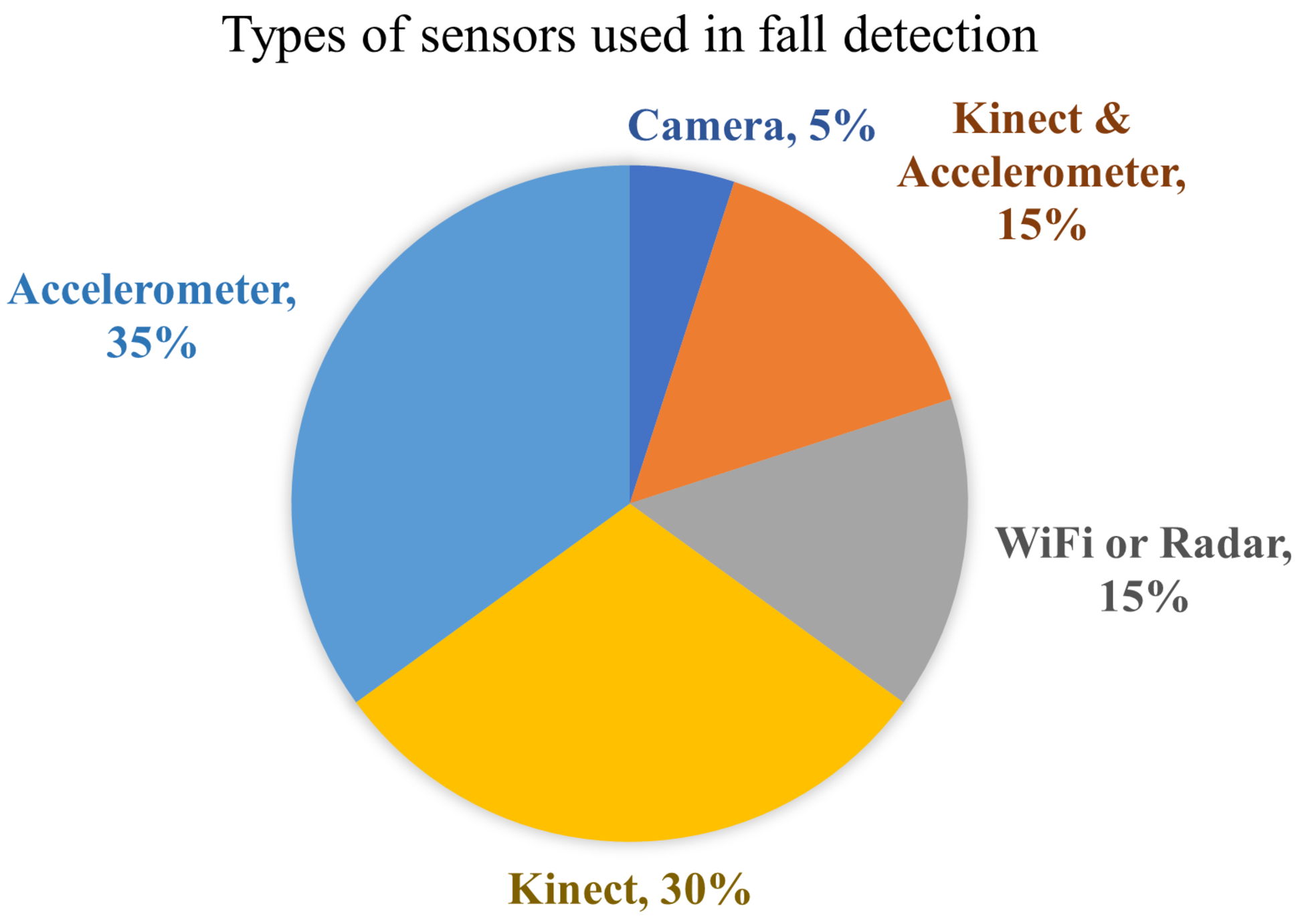}
\caption{Percentage of different sensor types used in fall detection systems from 2014 to 2018. “Kinect \& accelerometer” represents systems using both sensors \cite{Tao_2018}.}
\label{fig:sensor_types}
\end{figure}

Fig.~\ref{fig:sensor_types} shows the different types of sensors used in fall detection systems from 2014 to 2018. It is evident that the use of Kinect sensors in vision-based monitoring quickly gained popularity after its introduction while the use of accelerometer-based features has dropped from 55\% to 35\% \cite{Tao_2018}.

\subsection{Wearable sensors}
This technology encompasses different types of sensors, including inclinometers, gyroscopes or accelerometers. Most methods use one or a combination of sensor inputs followed by a threshold technique to detect fall. Nahian et al. proposed a feature selection method based on Pearson correlation coefficient to discard less dominant features from the multivariate accelerometer signal \cite{Nahian_access}. Alarifi et al. used principal component analysis to select more relevant features from a set of hand crafted features \cite{Alarifi_2021}. Then killer heuristic optimized AlexNet is used to recognize fall patters.

Although these methods have an advantage of low cost and follow the person where they are, they also have the inconvenience that the device is to be worn at all times which can be uncomfortable. Elderly people may easily forget to wear the device, making it ineffective for practical applications. Besides, these methods do not capture distinct patterns in the motion trajectories of activities. As a result, daily movements like shaking an object or sudden changes in posture can easily create false alarms. Hence, these methods suffer from high false positive rate.

%\subsection{Environmental sensors}

\subsection{Vision based techniques}
Lin et al. \cite{Lin_lstm} implemented an RNN architecture that takes skeleton data as input extracted from video frames using OpenPose. Keskes et al. \cite{Keskes_2021} used ST-GCN model \cite{Sijie_Yan_2018_aaai} to detect fall from skeleton body joints. They first trained the model on NTU-RGB+D dataset \cite{ShahroudyAmir2016NRAL} for action recognition then used the pre-trained model for fall detection. Carlier et al. \cite{Carlier_2020_EMBC} proposed a model that processes dense optical flow calculated from RGB videos using VGG-16 architecture. Optical flow, though quite effective in representing motion change, is very expensive to calculate. Wang et al. \cite{Wang_2020} used OpenPose to calculate body key points and subsequently calculated a set of dynamic and static features followed by MLP and Random Forest for classification. Though they achieved superior results than \cite{Carlier_2020_EMBC}, it is difficult to create a generalized model with handcrafted features.

Privacy is an important issue while developing a system to monitor daily lives. To conceal direct video feed, Liu et al. \cite{Liu_ICAIBD} proposed a multi-layer compressed sensing model that encrypts the video frames into a visually unintelligible form. Local binary patterns are then calculated to extract object's behaviour. They achieved good accuracy but on small datasets.%, it is possible to reverse engineer the encryption process to get the original video which poses a security breach in case of a leak.

Feng et al. proposed an attention guided method to detect fall in crowded places \cite{FENG2020242}. Their method first detects pedestrians using YOLOv3 model and then the detected region is processed through a VGG16 network to get their features. The output features are passed to an LSTM network to extract temporal information. Hence, they used VGG16 as a spatial feature extractor and LSTM as a temporal feature extractor. Xu et al. used Inception-ResNet-v2 to process 2D skeleton data extracted from frames using OpenPose \cite{XU2020123205}. Cai et al. proposed a multi-task hourglass convolutional autoencoder that learns to classify fall events (main task) and frame reconstruction (auxiliary task) \cite{Cai_AE_2020}.

Overall from the literature, we found that most of the earlier work on wearable and environmental sensor data focused on handcrafted features and classical machine learning methods. A major limitation of these techniques is that handcrafted features do not perfectly reveal the inherent pattern of complex action sequences. On the other hand, vision-based methods apply learning techniques to extract motion structures and they mainly use image processing models. Besides, these studies report their model evaluation on very small datasets containing only a small set of daily activity classes. Therefore, their efficiency in real-life scenarios is not comprehensible. 

Recently graph convolutional networks (GCN) have gained tremendous success in human action recognition \cite{Sijie_Yan_2018_aaai, liu2020disentangling, song2020stronger}. These networks can extract rich spatio-temporal features directly from skeleton data. %Comprehensive studies have not been done in this area for fall detection systems. 
Therefore, in this work, we propose a novel method using the effectiveness of GCN for fall detection. We tested our proposed method on the largest available datasets. %Our proposed methodology is explained in detail in Section~\ref{method}. 

\section{Proposed Skeleton-Based Fall Detection}\label{method}
%Earlier approaches \cite{Nahian_access} to fall detection mostly focused on handcrafted features for classical machine learning methods which ignore the important spatio-temporal information in human motion. Therefore, these methods which often used a threshold technique to differentiate fall event can give high false rate for fall like activities. 
Recently different approaches \cite{Carlier_2020_EMBC, Wang_2020, Liu_ICAIBD, FENG2020242, XU2020123205, Cai_AE_2020} have been proposed which significantly improved the overall fall detection accuracy. However, these methods use RGB videos or skeleton data extracted from RGB frames and processed them with traditional image recognition models such as InceptionResNet or VGG16. Though skeleton data are advantageous in many aspects compared to videos, these models cannot fully exploit the semantic relationship among human joints as they treat all joints independently without acknowledging their internal connectivity. Besides, joints form local groups for a certain activity and the relative locations of these groups are important factors. Joint connectivity can assist in tracking these local and global spatio-temporal trajectories. 

\begin{figure*}[htbp]
\centering
\includegraphics[width=18cm,height=3cm]{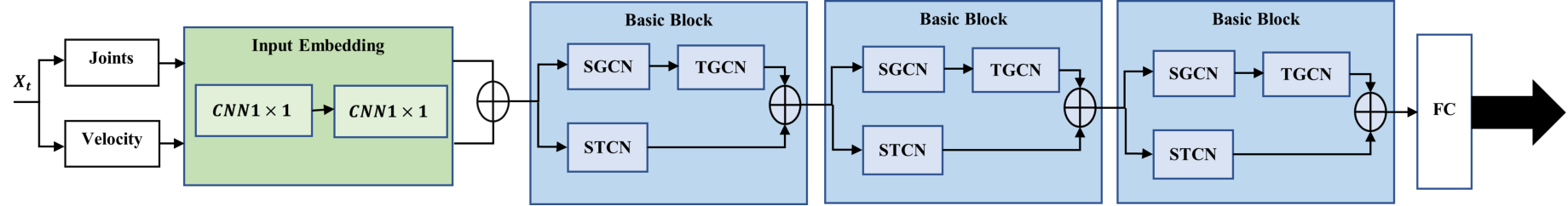}
\caption{Architecture overview: Input embedding has a two-layer 1D-CNN block that normalizes the frames and creates an enhanced projection, then both the embedded joint and velocity streams are concatenated. Each basic block deploys a multi-pathway of spatial-temporal and Conv2D modules to capture spatio-temporal dependencies.}
\label{fig:proposed_model}
\end{figure*}

Graph convolution uses adjacency matrix that incorporates node connectivity information to the convolution operation to extract inherent dependency. Human skeleton can also be represented as a graph with the joints being nodes and the bones as edges. Therefore, we propose a GCN based architecture that utilizes physical and learnable node adjacency to learn better body structure representation. Fig.~\ref{fig:proposed_model} illustrates our proposed methodology. It has an input embedding layer followed by three basic block layers and final fully connected layer for classification. Each basic block deploys two types of pathways to simultaneously capture complex spatio-temporal correlations. The SGCN-TGCN pathway constructs a spatial-temporal window, performing spatial convolution then temporal convolution to extract rich spatio-temporal dependency. STCN extends the representativeness of the network to capture features with expanding receptive fields.

\subsection{Input Embedding}
Skeleton data is captured as a sequence of frames, each frame containing a set of 2D or 3D joints. The input feature set $x_{c\times t \times v}$ represents a sample with $c$ number of channels with $v$ nodes at time $t$. Since the speed of change in joint positions is the distinct characteristic of fall event, we create a dynamic representation of the skeleton frames by adding joint information and velocity information together. Joint velocity $v_t$ is calculated as Eq~\ref{joint_velocity}.

\begin{equation}\label{joint_velocity}
v_t = x_{(t_i)} - x_{(t_{i-1})}
\end{equation}
where $x{(t_i)}$ is joint locations at time step $t_i$ and $x{(t_{i-1})}$ is joint locations at previous time step $t_{i-1}$. Both joint and velocity are encoded through the embedding block. It contains a batch normalization layer and two $1\times1$ projection layers each followed by ReLU activation function \cite{relu_icml_2010} as:

\begin{equation}\label{embedding_eq}
\widehat{x_t} = \sigma(W_2(\sigma(W_1x_t+b1))+b2)
\end{equation}
where $W_1,W_2 \in \mathbb{R}^{c\times c}$ are weights for $c$ number of channels, $b1$ and $b2$ are bias matrices. $\sigma$ represents ReLU activation function. Due to subtraction, velocity matrix $v_t$ has one less frame. Therefore, we concatenate an all-zero frame in the beginning representing no motion change in the first frame. Then it is also embedded using the projection layers following Eq~\ref{embedding_eq}. Finally, the projected joint and velocity are combined together by summation as follows.

\begin{equation}\label{z_t}
Z_t = \widehat{x_t} + \widehat{v_t}
\end{equation}

\subsection{Skeleton Graph Representation}
\begin{figure}[htbp]
\centering
\includegraphics[height=5cm, width=8.5cm]{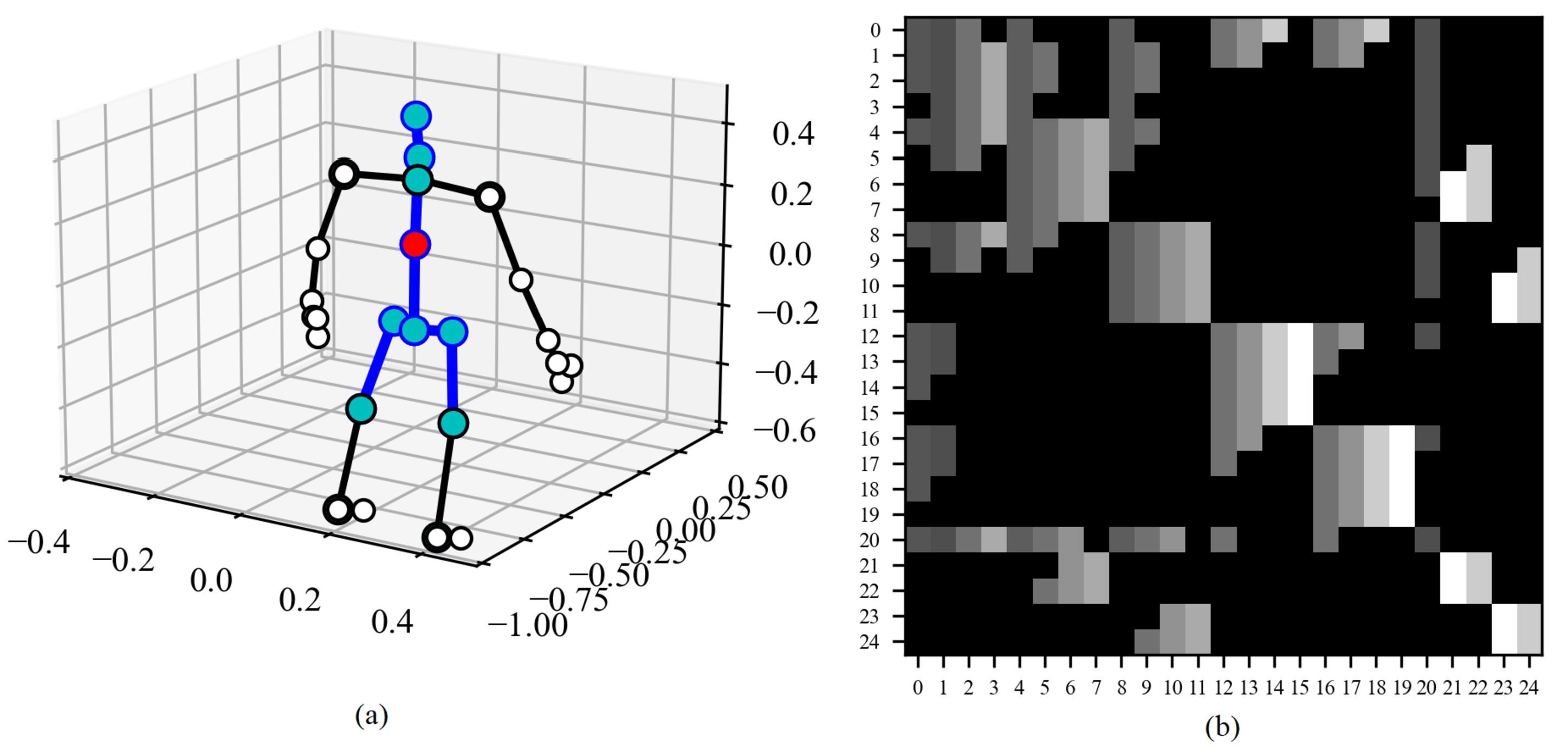}
\caption{Illustration of joint adjacency with 3 hop distance. \textbf{(a):} skeleton representation of neighbors (blue) of spine joint (red). \textbf{(b):} $N\times N$ matrix representation for all 25 joints.}
\label{fig:skeleton_hop_adjacency}
\end{figure}

A human skeleton is denoted as $g=(v,\varepsilon)$, where $v$ is the set of $N$ nodes representing joints and $\varepsilon$ is the set of edges representing physical connections. Latent connectivity of the skeleton graph is represented with a $H$ hop adjacency matrix $A\in{R^ {N\times N}}$ where initially $A_{i,j}=1$ for direct physical connection between $v_i$ and $v_j$, and nodes at maximum $H$ edge distance to include neighboring nodes into account and 0 otherwise. So node affinity is extended beyond self loop and direct connection. Finally, the adjacency matrix is normalized. Besides the adjacency matrix, a learnable parameter is added at each layer to capture data adaptive joint relevance information during training. Since the static adjacency matrix only represents visual connectivities, it provides same information for all convolutional layers. This added learnable parameter can grasp latent association at higher level. Hence, the adjacency matrix becomes $A\times \theta^l$ where $\theta^l$ represents learnable edge importance parameter at layer $l$. Fig. \ref{fig:skeleton_hop_adjacency} shows a skeleton representation with 3 hop edge adjacency. 

\subsection{Spatial Graph Convolution Network (SGCN)}
Traditional 2D convolution treats input as a 2D image and calculates output feature map using the specified filter to capture shapes and edges. Adjacency matrix acts like a message passing operation enabling graph convolution to learn node adjacency. We used the spatial configuration partitioning for generating adjacency matrix suggested by Yan et al. in \cite{Sijie_Yan_2018_aaai} where neighboring nodes are divided into three subsets: root node, centripetal group: closer to the gravity center of root node, and centrifugal group: farther to root node. Instead of using the usual $N\times N$ adjacency, this distance partitioning enhances graph representativeness for spatial convolution. SGCN can be defined as:

\begin{equation}\label{sgcn_eq}
f_s(Z_t) = \sigma(\hat{A}*Z_t*W+b)
\end{equation}
where $\hat{A} = A*\theta^l$  is the adjacency matrix with learnable weight accumulation $W$ at layer $l$. The term $\hat{A}*Z_t*W$ can be interpreted as the spatial feature aggregation of $Z_t$, the dynamic embedded feature from Eq~\ref{z_t}, from close and far apart neighboring nodes. The aggregated feature farther enhanced with a residual connection is processed through an activation layer.

\subsection{Temporal Graph Convolution Network(TGCN)}
After spatial aggregation through SGCN, we apply temporal graph convolution to model the temporal dynamics in the skeleton frames. Same joints across the entire sequence are connected together. A temporal window filter is applied to aggregate features over time. Thus, the neighborhood concept is extended both spatially and temporally.

\subsection{Joint and frame level Convolution Network(STCN)}
The disjoint SGCN and TGCN applies spatial and temporal convolution separately. Thus the complex spatio-temporal joint correlation is less powerful. It is evident that there is a strong connection among joints in a single frame and their transition over time. This relationship is further weakened in the higher layers with increasingly larger spatial and temporal receptive field. Therefore, we extend the representativeness of the feature aggregation of the independent graph convolution operations by introducing STCN block. It applies a spatio-temporal kernel to exploit the correlation across both joints and frames. For instance, take two similar actions such as \emph{Fall} and \emph{Lying down}, the spatial and temporal configurations are comparatively similar if considered separately. But intuitively the weighting should be different as the spatial change of the joint coordinates over time for \emph{Fall} happens far too quickly than \emph{Lying down}. STCN achieves this with the unified spatio-temporal feature aggregator.

\section{Experiments}\label{results}
\subsection{Datasets}\label{datasets}
\textbf{UWA 3D Multiview Activity II}: UWA 3D Multiview Activity II dataset \cite{Rahmani_tpami_2016} contains 1015 samples collected from 10 subjects %Ajmal: 20 subjects??
with 4 different viewpoints at different times. Each subject performed thirty different daily activities and repeated them four times, including fall and lying down on floor. While most existing datasets use different camera at the same time to capture different viewpoint, here activities are re-performed for each viewpoint. Hence it poses a greater challenge since the same subject performing the same activity at different times will produce different action patterns. As a result, this dataset provides a more diversified viewpoint scenario. We used two evaluation settings as suggested in \cite{Rahmani_tpami_2016}: (1) \textit{Cross-View: Val3}, where camera 1 and 2 samples are used for training and camera 3 for testing, (2) \textit{Cross-View: Val4}, where camera 1 and 2 samples are used for training and camera 4 for testing. 

\textbf{NTU 60}: NTU 60 \cite{ShahroudyAmir2016NRAL} is a large scale dataset containing 60 different daily life activity classes including \emph{Fall} and similar actions such as squatting down. It has 56,578 skeleton sequences captured from 40 different subjects with 3 camera views. Each sequence contains 25 body joints in 3D coordinates with 1 or 2 subjects. Among these, 1949 are \emph{Fall} samples. We used two evaluation settings as suggested in \cite{ShahroudyAmir2016NRAL}: (1) \textit{Cross-Subject}, where subjects are divided into training and testing splits, yielding 40,091 and 16,487 training and testing examples respectively, (2) \textit{Cross-View}, where camera 2 and 3 samples are used for training and rest for testing. 

\textbf{NTU 120}: NTU 120 \cite{Liu_2019_NTURGBD120} extends NTU 60 with another 60 action classes and an additional 57,367 skeleton sequences, totalling 113,945 samples over 120 classes captured from 106 distinct subjects and 32 camera view setup. Among these, 2017 samples are \emph{Fall} samples. We used two evaluation settings as suggested in \cite{Liu_2019_NTURGBD120}: (1) \textit{Cross-Subject}, same as NTU 60 and, (2) \textit{Cross-Setup}, where even setups are used for training and rest for testing.

\begin{figure*}[htbp]
\centering
\includegraphics[height=12cm, width=16cm]{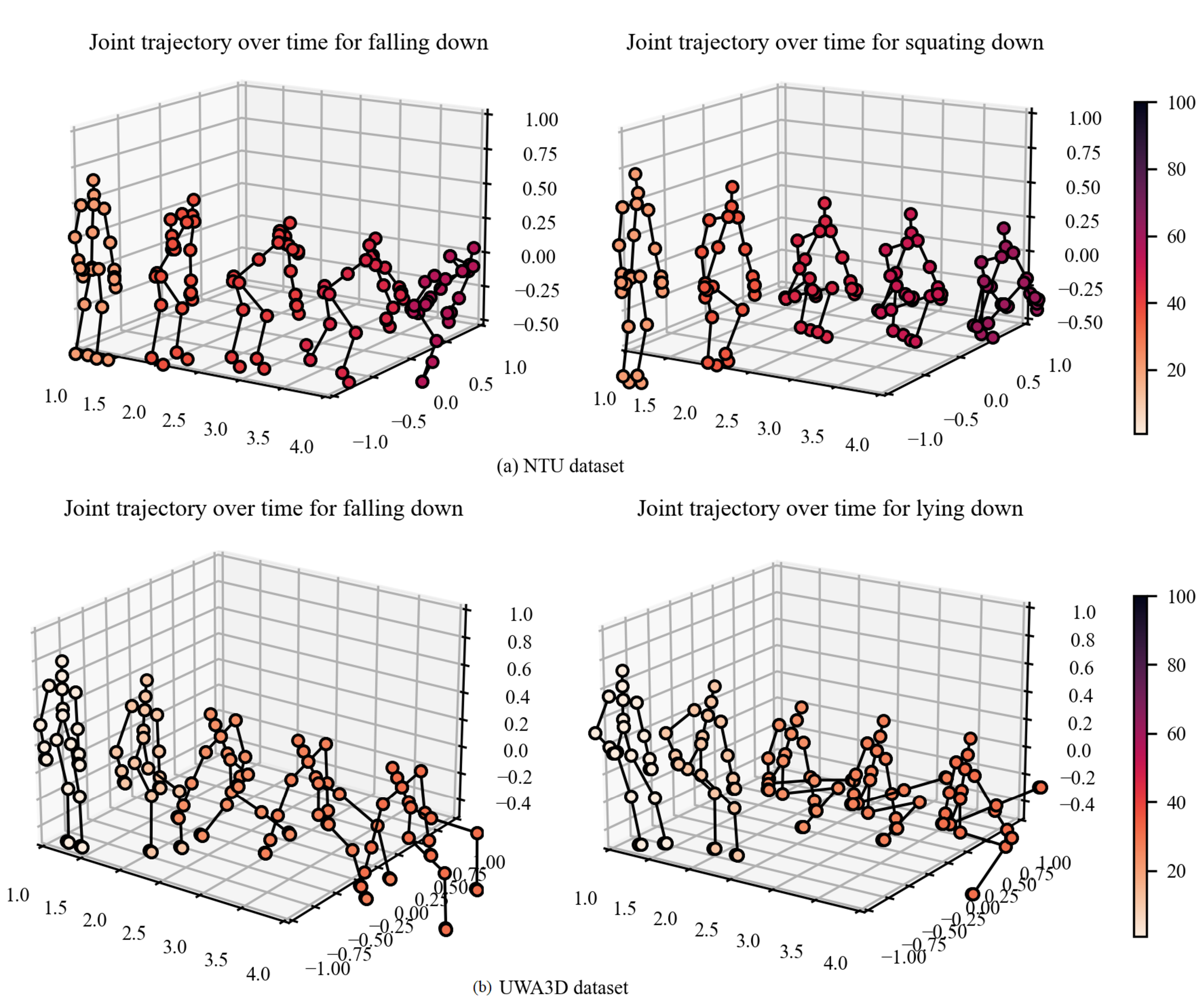}
\caption{Skeleton joint trajectory of two activities with similar posture over time in NTU (top) and UWA3D (bottom) datasets.} 
\label{fig:joint_trajectory_ntu}
\end{figure*}

Fig.~\ref{fig:joint_trajectory_ntu} shows two \emph{Fall} and \emph{Non-Fall} activity samples from NTU and UWA3d datasets. Both samples have very similar posture especially the \emph{Lying down} event in UWA3D dataset has much closer resemblance to \emph{Fall} event considering the posture change over time. 

For our fall detection experiments, we considered \emph {Fall} samples as positive class and all other samples as negative. To account for the imbalance in the dataset, we used a balanced batch of approximately 50-50 positive-negative samples at each iteration. The two NTU datasets \cite{ShahroudyAmir2016NRAL,Liu_2019_NTURGBD120} are particularly difficult than all the other available fall datasets as they contain a large collection of daily life activity classes. The datasets closely resemble natural activities happening around in a household and provide a more robust way of evaluation for real-life implementation.

\subsection{Implementation Details}
Since activity sequences have different number of frames, we replayed them to have a maximum of 300 frames. Frames that do not contain any skeleton are discarded and translations are performed following \cite{lei_shi_2019_2s_agcn} to achieve view-invariant transformations by centring the skeleton at the spine joint and paralleling the shoulder bone to the x-axis and the bone between hip and spine to the z-axis. We used a temporal window of size 250 to extract frames at a random starting point so that the model learns to detect falls at different starting physical postures. Joint coordinates are normalized to have zero mean and unit variance.

We trained our model for 30 epochs using SGD with a batch size 64 (32 per worker), 0.9 momentum and an initial learning rate of 0.05 decreasing by 10\% after every 10 epochs. Weight decay was set to 0.0005. These hyperparameters are fixed for all the datasets across all evaluation settings.

\subsection{Results}
Experimental results are presented in Table \ref{tab:experimental_results}. We report our results in terms of F1 score, sensitivity/recall, specificity, Area under the ROC Curve (AUC), False Positive rate (FP rate) and accuracy. Traditional random train-test split (70-30\%) does not account for person and view point variations which is essential for reliable performance measurement. Therefore, we followed standard practice of cross-subject (X-Sub) and cross-view/setup (X-View/X-Set) evaluation protocol used in action recognition, to evaluate our model.

\begin{table*}[htbp]
\small
  \fontsize{9pt}{10pt}
  \centering   
  \caption{Experimental results of our proposed fall detection method. X-Sub means cross subject, X-View means cross view and X-Set means cross set experimental setting. }% without any pre-training or finetuning. Evaluation settings are explained in Section~\ref{datasets}. We achieved excellent results with very low false positive rates.}
    \begin{tabular}[1pt]{|p{2.5cm}|p{1.5cm}|p{1.3cm}|p{1.5cm}|p{1.5cm}|p{1cm}|p{1.1cm}|p{1.5cm}|}
    \hline
    \makecell{\textbf{Dataset}} & \makecell{\textbf{Evaluation}} &
    \makecell{\textbf{F1 Score}} & \makecell{\textbf{Sensitivity}} & \makecell{\textbf{Specificity}} & \makecell{\textbf{AUC}} & \makecell{\textbf{FP rate}} & \makecell{\textbf{Accuracy}} \bigstrut\\
    \hline
    UWA3D & \hfil Val3 & \hfil 75.00\% & \hfil 66.67\%& \hfil 99.59\%& \hfil 83.13\%& \hfil 0.42\%& \hfil 98.40\%\bigstrut\\
    \hline
    UWA3D & \hfil Val4 & \hfil 75.00\% & \hfil 66.67\%& \hfil 99.59\%& \hfil 83.13\%& \hfil 0.41\%& \hfil 98.43\%\bigstrut\\
    \hline
    NTU 60 & \hfil X-Sub & \hfil 96.00\% & \hfil 96.00\%& \hfil 99.93\%& \hfil 97.97\%& \hfil 0.07\%& \hfil 99.86\%\bigstrut\\
    \hline
    NTU 60 & \hfil X-View & \hfil 97.79\%& \hfil 98.10\%& \hfil 99.96\%& \hfil 99.03\%& \hfil 0.04\%& \hfil 99.93\%\bigstrut\\
    \hline
    NTU 120 & \hfil X-Sub & \hfil 85.38\% & \hfil 92.36\%& \hfil 99.87\%& \hfil 96.12\%& \hfil 0.13\%& \hfil 99.83\%\bigstrut\\
    \hline
    NTU 120 & \hfil X-Set & \hfil 90.95\% & \hfil 91.67\%& \hfil 99.92\%& \hfil 95.79\%& \hfil 0.09\%& \hfil 99.85\%\bigstrut\\
    \hline
    
    \end{tabular}%
  \label{tab:experimental_results}%
\end{table*}%

Retraining a model repeatedly, to account for new types of daily activities, is a time consuming and impractical solution. Such a situation commonly arises in a household with the arrival of new people or simply as a result of numerous new daily activities. Therefore, we tested our pretrained model on other datasets without fine-tuning the model. Table \ref{tab:transfer_learning} shows results of our model trained on the NTU 60 dataset and tested on other datasets. Though NTU 120 contains 60 new activity classes, our model is able to identify fall events with high accuracy. Additionally, UWA3D dataset contains an activity \emph{Lying down on the floor} which has very similar progression of postures like fall with the subtle difference in speed of change. Our model effectively distinguished these sequences from fall. In addition, on UWA3D the samples from different camera viewpoint are captured separately. The fact that our model performed well on both evaluation for UWA3D without any finetuning demonstrates the effectiveness of the viewpoint invariant feature extraction capability  as well as the generalization capability of our model.

\begin{table}[htbp]
\small
    \fontsize{9pt}{10pt}
  \centering   
  \caption{Accuracy of fall detection when our model was trained on one dataset and tested on another dataset with unseen activities and subjects.}
    \begin{tabular}[1pt]{|p{1.85cm}|p{2.3cm}|p{1.5cm}|p{1.3cm}|}
    \hline
    \makecell{\textbf{Traning }} & \makecell{\textbf{Test dataset}} & \makecell{\textbf{Evaluation}} & \makecell{\textbf{Accuracy}} \bigstrut\\
    \makecell{\textbf{dataset}} & \makecell{\textbf{}} & \makecell{\textbf{protocol}} & \makecell{\textbf{}} \bigstrut\\
    \hline
    \multirow{4}{*}{NTU 60} & \multirow{2}{*}{NTU 120} &  \hfil X-Sub & \hfil 98.49\% \bigstrut\\ \cline{3-4} 
                  &                 & \hfil X-Set & \hfil 98.90\% \bigstrut\\ \cline{2-4} 
                  & \multirow{2}{*}{UWA3D} & \hfil Val3 & \hfil  88.00\% \bigstrut\\ \cline{3-4} 
                  &                   & \hfil Val4 & \hfil 92.91\% \bigstrut\\  \hline
    \end{tabular}
  \label{tab:transfer_learning}%
\end{table}

\subsection{Comparison with state of the art methods}
We compare our model with the state-of-the-art methods on NTU 60 dataset in Table \ref{tab:ntu_sota}. There are only three works that reported results on NTU 60 dataset: Inception-ResNet-v2 \cite{XU2020123205}, ST-GCN \cite{Keskes_2021} and 1D-CNN \cite{Tsai_2019}. Xu et al. \cite{XU2020123205} did not specify their evaluation metric and both Keskes et al. \cite{Keskes_2021} and Tsai et al. \cite{Tsai_2019} used X-Sub evaluation protocol only. Our model with a much simpler architecture and no data augmentation achieved superior performance under all evaluation protocols.  

\begin{table}[htbp]
\small
  \fontsize{9pt}{10pt}
  \centering   
  \caption{Fall detection accuracy comparison against state-of-the-art methods on NTU 60 dataset. We report accuracy results on both cross-subject (X-Sub) and cross-view (X-View) evaluation settings.}
    \begin{tabular}[1pt]{|p{2.2cm}|p{1.5cm}|p{1.9cm}|p{1.3cm}|}
    \hline
    \makecell{\textbf{Methods}} & \makecell{\textbf{Modality}} & 
    \makecell{\textbf{Evaluation}} & \makecell{\textbf{Accuracy}} \bigstrut\\
    \hline
     Inception-ResNet-v2\cite{XU2020123205}& \hfil RGB & Not mentioned & \hfil 91.70\% \bigstrut\\
    \hline
     ST-GCN \cite{Keskes_2021} & \hfil Skeleton & \hfil X-Sub & \hfil 92.90\% \bigstrut\\
    \hline
     1D-CNN \cite{Tsai_2019} & \hfil Skeleton & \hfil 70\%-30\% & \hfil 99.20\% \bigstrut\\
    \hline
     \multirow{2}{*}{\textbf{Proposed}} & \hfil \multirow{2}{*}{Skeleton} & \hfil X-Sub & \hfil \textbf{99.86\%} \bigstrut\\\cline{3-4} 
     &  & \hfil X-View & \hfil \textbf{99.93\%} \bigstrut\\
    \hline
    \end{tabular}%
  \label{tab:ntu_sota}%
\end{table}%

Table \ref{tab:architecture_comparison} presents the comparison between our proposed architecture and the current state-of-the-art model \cite{Keskes_2021} in terms of the number of model parameters, training time and inference time. We did not consider \cite{XU2020123205} as they used RGB data and could not compare these parameters with \cite{Tsai_2019} as there is no published code. Our model exceeds in all measures with a significant improvement with an order of magnitude smaller architecture and achieves 7.5 times faster average inference time.

\begin{table}[htbp]
\small
  \fontsize{9pt}{10pt}
  \centering   
  \caption{Comparison of our model with the state-of-the-art on NTU 60 in terms of the number of parameters, FLOPS, training and inference time.}
    \begin{tabular}[1pt]{|p{1.8cm}|p{1.4cm}|p{1cm}|p{1.7cm}|p{1.21cm}|}
    \hline
    \makecell{\textbf{Methods}} & \makecell{\textbf{\thead{No. of\\parameters}}} & \makecell{\textbf{\thead{FLOPS}}} &
    \makecell{\textbf{\thead{Training time\\(min/epoch)}}} &
    \makecell{\textbf{\thead{Average\\Inference\\time (ms)}}} \bigstrut\\
    \hline
     ST-GCN \cite{Keskes_2021} & \hfil 4.38M & \hfil 24.59G & \hfil 20.18 & \hfil 9.95 \bigstrut\\
    \hline
     \textbf{Proposed} & \hfil \textbf{1.26M} &\hfil \textbf{16.17G} & \hfil  \textbf{3.23} & \hfil \textbf{1.32} \bigstrut\\
    \hline
    \end{tabular}%
  \label{tab:architecture_comparison}%
\end{table}%

%\section{Discussion}\label{discussion}

\section{Conclusion}\label{conclusion}
In this work, we presented an effective fall detection method that has two advantages: it uses a skeleton based approach to eliminate privacy concerns, and it has a lightweight architecture which is easier to implement on embedded systems. Our proposed technique applies a robust computing block that can capture discriminative spatio-temporal co-occurrence relationships of joints over time in a two-stream fashion. The use of the separate streams in the microcomputing block gives our method a powerful feature extraction capability without burdening the overall network architecture. We used a smaller window to extract sequence of frames from a random starting point. This shows the model's superior capability to detect fall at any stage in a household where occlusion is quite common due to furnitures. With extensive experiments on largescale datasets, we showed that our model outperforms existing methods and achieves state-of-the-art results.


\section*{Acknowledgment}
This research was supported by ARC DP190102443.



%\bibliographystyle{IEEEtran}
\Urlmuskip=0mu plus 1mu\relax
\bibliographystyle{IEEEtran}
\bibliography{bibtex}

\begin{thebibliography}{10}
\providecommand{\url}[1]{#1}
\csname url@samestyle\endcsname
\providecommand{\newblock}{\relax}
\providecommand{\bibinfo}[2]{#2}
\providecommand{\BIBentrySTDinterwordspacing}{\spaceskip=0pt\relax}
\providecommand{\BIBentryALTinterwordstretchfactor}{4}
\providecommand{\BIBentryALTinterwordspacing}{\spaceskip=\fontdimen2\font plus
\BIBentryALTinterwordstretchfactor\fontdimen3\font minus
  \fontdimen4\font\relax}
\providecommand{\BIBforeignlanguage}[2]{{%
\expandafter\ifx\csname l@#1\endcsname\relax
\typeout{** WARNING: IEEEtran.bst: No hyphenation pattern has been}%
\typeout{** loaded for the language `#1'. Using the pattern for}%
\typeout{** the default language instead.}%
\else
\language=\csname l@#1\endcsname
\fi
#2}}
\providecommand{\BIBdecl}{\relax}
\BIBdecl

\bibitem{WHO_FALL}
``World health organization - falls,''
  \url{www.who.int/news-room/fact-sheets/detail/falls}, 2021, [Online; accessed
  28-June-2021].

\bibitem{AIHW_FALL}
``Injury in {A}ustralia: falls,'' \url{www.aihw.gov.au/reports/injury/falls},
  2021, [Online; accessed 29-June-2021].

\bibitem{ANZ_FALL}
``Australian and new zealand falls prevention society - info about falls,''
  \url{www.anzfallsprevention.org/info/}, 2021, [Online; accessed
  28-June-2021].

\bibitem{UN_age}
``Department of {E}conomic and {S}ocial {A}ffairs, {U}nited {N}ations,''
  \url{www.un.org/en/development/desa/population/publications/pdf/ageing/WPA2017\_Highlights.pdf},
  2017, [Online; accessed 28-June-2021].

\bibitem{long_lie_jama}
M.~E. Tinetti, W.-L. Liu, and E.~B. Claus, ``{Predictors and Prognosis of
  Inability to Get Up After Falls Among Elderly Persons},'' \emph{JAMA}, 1993.

\bibitem{MATHON2017e50}
C.~Mathon, F.~Beaucamp, F.~Roca, P.~Chassagne, A.~Thevenon, and F.~Puisieux,
  ``Post-fall syndrome: Profile and outcomes,'' \emph{Annals of Physical and
  Rehabilitation Medicine}, 2017, 32nd Annual Congress of the French Society of
  Physical and Rehabilitation Medicine.

\bibitem{NYAN_2008}
M.~Nyan, F.~E. Tay, and E.~Murugasu, ``A wearable system for pre-impact fall
  detection,'' \emph{Journal of Biomechanics}, 2008.

\bibitem{Fakhrulddin_2017}
A.~H. Fakhrulddin, X.~Fei, and H.~Li, ``Convolutional neural networks ({CNN})
  based human fall detection on body sensor networks ({BSN}) sensor data,'' in
  \emph{2017 4th International Conference on Systems and Informatics (ICSAI)},
  2017.

\bibitem{NGUYENGIA_2018}
T.~{Nguyen Gia}, V.~K. Sarker, I.~Tcarenko, A.~M. Rahmani, T.~Westerlund,
  P.~Liljeberg, and H.~Tenhunen, ``Energy efficient wearable sensor node for
  iot-based fall detection systems,'' \emph{Microprocessors and Microsystems},
  2018.

\bibitem{Alarifi_2021}
A.~Alarifi and A.~Alwadain, ``Killer heuristic optimized convolution neural
  network-based fall detection with wearable iot sensor devices,''
  \emph{Measurement}, vol. 167, 2021.

\bibitem{Anceschi_2021}
E.~Anceschi, G.~Bonifazi, M.~C. De~Donato, E.~Corradini, D.~Ursino, and
  L.~Virgili, \emph{SaveMeNow.AI: A Machine Learning Based Wearable Device for
  Fall Detection in a Workplace}.\hskip 1em plus 0.5em minus 0.4em\relax
  Springer International Publishing, 2021.

\bibitem{Lin_lstm}
C.-B. Lin, Z.~Dong, W.-K. Kuan, and Y.-F. Huang, ``A framework for fall
  detection based on openpose skeleton and {LSTM/GRU} models,'' \emph{Applied
  Sciences}, 2021.

\bibitem{Keskes_2021}
O.~Keskes and R.~Noumeir, ``Vision-based fall detection using {ST-GCN},''
  \emph{IEEE Access}, 2021.

\bibitem{Carlier_2020_EMBC}
A.~Carlier, P.~Peyramaure, K.~Favre, and M.~Pressigout, ``Fall detector adapted
  to nursing home needs through an optical-flow based {CNN},'' in \emph{42nd
  Annual International Conference of the IEEE Engineering in Medicine \&
  Biology Society (EMBC)}, July 2020.

\bibitem{Wang_2020}
B.-H. Wang, J.~Yu, K.~Wang, X.-Y. Bao, and K.-M. Mao, ``Fall detection based on
  dual-channel feature integration,'' \emph{IEEE Access}, 2020.

\bibitem{Liu_ICAIBD}
J.-x. Liu, R.~Tan, N.~Sun, G.~Han, and X.-f. Li, ``Fall detection under privacy
  protection using multi-layer compressed sensing,'' in \emph{2020 3rd
  International Conference on Artificial Intelligence and Big Data (ICAIBD)},
  2020.

\bibitem{FENG2020242}
Q.~Feng, C.~Gao, L.~Wang, Y.~Zhao, T.~Song, and Q.~Li, ``Spatio-temporal fall
  event detection in complex scenes using attention guided {LSTM},''
  \emph{Pattern Recognition Letters}, 2020.

\bibitem{XU2020123205}
Q.~Xu, G.~Huang, M.~Yu, and Y.~Guo, ``Fall prediction based on key points of
  human bones,'' \emph{Physica A: Statistical Mechanics and its Applications},
  2020.

\bibitem{Cai_AE_2020}
X.~Cai, S.~Li, X.~Liu, and G.~Han, ``Vision-based fall detection with
  multi-task hourglass convolutional auto-encoder,'' \emph{IEEE Access}, 2020.

\bibitem{Gunnar_1973}
G.~Johansson, ``Visual perception of biological motion and a model for its
  analysis,'' \emph{Perception \& Psychophysics}, 1973.

\bibitem{Liu_2019_NTURGBD120}
J.~Liu, A.~Shahroudy, M.~Perez, G.~Wang, L.-Y. Duan, and A.~C. Kot, ``{NTU
  RGB+D 120}: A large-scale benchmark for 3d human activity understanding,''
  \emph{IEEE Transactions on Pattern Analysis and Machine Intelligence}, 2019.

\bibitem{Tao_2018}
T.~Xu, Y.~Zhou, and J.~Zhu, ``New advances and challenges of fall detection
  systems: A survey,'' \emph{Applied Sciences}, 2018.

\bibitem{Nahian_access}
M.~J.~A. Nahian, T.~Ghosh, M.~H.~A. Banna, M.~A. Aseeri, M.~N. Uddin, M.~R.
  Ahmed, M.~Mahmud, and M.~S. Kaiser, ``Towards an accelerometer-based elderly
  fall detection system using cross-disciplinary time series features,''
  \emph{IEEE Access}, 2021.

\bibitem{Sijie_Yan_2018_aaai}
S.~Yan, Y.~Xiong, and D.~Lin, ``Spatial temporal graph convolutional networks
  for skeleton-based action recognition,'' in \emph{Thirty-Second AAAI
  Conference on Artificial Intelligence}, February 2018.

\bibitem{ShahroudyAmir2016NRAL}
A.~Shahroudy, J.~Liu, T.-T. Ng, and G.~Wang, ``{NTU RGB+D}: A large scale
  dataset for 3d human activity analysis,'' in \emph{The IEEE Conference on
  CVPR}, June 2016.

\bibitem{liu2020disentangling}
Z.~Liu, H.~Zhang, Z.~Chen, Z.~Wang, and W.~Ouyang, ``Disentangling and unifying
  graph convolutions for skeleton-based action recognition,'' in
  \emph{Proceedings of the IEEE/CVF Conference on Computer Vision and Pattern
  Recognition}, 2020.

\bibitem{song2020stronger}
Y.-F. Song, Z.~Zhang, C.~Shan, and L.~Wang, ``Stronger, faster and more
  explainable: A graph convolutional baseline for skeleton-based action
  recognition,'' in \emph{Proceedings of the 28th ACM International Conference
  on Multimedia (ACMMM)}, 2020.

\bibitem{relu_icml_2010}
V.~Nair and G.~E. Hinton, ``Rectified linear units improve restricted boltzmann
  machines,'' in \emph{The International Conference on Machine Learning
  (ICML)}, 2010.

\bibitem{Rahmani_tpami_2016}
H.~Rahmani, A.~Mahmood, D.~Huynh, and A.~Mian, ``Histogram of oriented
  principal components for cross-view action recognition,'' \emph{IEEE
  Transactions on Pattern Analysis and Machine Intelligence}, 2016.

\bibitem{lei_shi_2019_2s_agcn}
L.~Shi, Y.~Zhang, J.~Cheng, and H.~Lu., ``Two-stream adaptive graph
  convolutional networks for skeleton-based action recognition,'' in \emph{The
  IEEE Conference on CVPR}, 2019.

\bibitem{Tsai_2019}
T.-H. Tsai and C.-W. Hsu, ``Implementation of fall detection system based on
  3{D} skeleton for deep learning technique,'' \emph{IEEE Access}, 2019.

\end{thebibliography}

\end{document}